\documentclass[9pt]{article}

\usepackage{spconf,amsmath,graphicx}

\usepackage{cite}
\usepackage{amssymb,amsfonts}
\usepackage{algorithmic}
\usepackage{textcomp}
\usepackage{xcolor}
\usepackage{stfloats}
\usepackage{algorithm}

\usepackage{booktabs}  
\usepackage{threeparttable}
\usepackage{multirow}

\title{EE-AE: An Exclusivity Enhanced \\
Unsupervised Feature Learning Approach}
\name{Jingcai Guo, Song Guo\thanks{This work is supported by National Natural Science Foundation of China (61872310) and INTPART BDEM project.}}
\address{Department of Computing, The Hong Kong Polytechnic University, Hong Kong\\
\texttt{cscjguo@comp.polyu.edu.hk, song.guo@polyu.edu.hk}}
\begin{document}
%
\maketitle
\begin{abstract}
Unsupervised learning is becoming more and more important recently. As one of its key components, the autoencoder (AE) aims to learn a latent feature representation of data which is more robust and discriminative. However, most AE based methods only focus on the reconstruction within the encoder-decoder phase, which ignores the inherent relation of data, i.e., statistical and geometrical dependence, and easily causes overfitting. In order to deal with this issue, we propose an Exclusivity Enhanced (EE) unsupervised feature learning approach to improve the conventional AE. To the best of our knowledge, our research is the first to utilize such exclusivity concept to cooperate with feature extraction within AE. Moreover, in this paper we also make some improvements to the stacked AE structure especially for the connection of different layers from decoders, this could be regarded as a weight initialization trial. The experimental results show that our proposed approach can achieve remarkable performance compared with other related methods.
\end{abstract}
\section{Introduction and Related Work}
With the development of deep learning techniques, more and more applications adopt the deep neural networks (DNNs) to handle multiple tasks and obtain successive state-of-the-art performances \cite{lecun2015deep,anderson2018bottom}. However, most of such DNNs are categorized into the supervised learning \cite{wolpert2018status} with high demands for labeled data. On the other hand, it is much easier to collect the unlabeled data, then how to utilize it falls into unsupervised learning \cite{zhou2017unsupervised}. To this issues, the natural idea is to combine them. We apply the DNNs with the fashion of unsupervised learning, and aim to obtain more robust and discriminative feature representation without the rich resource of labeled data.
\begin{figure}[t]
	\centering
	\begin{minipage}{0.3\linewidth}
		\centerline{\includegraphics[width=3cm]{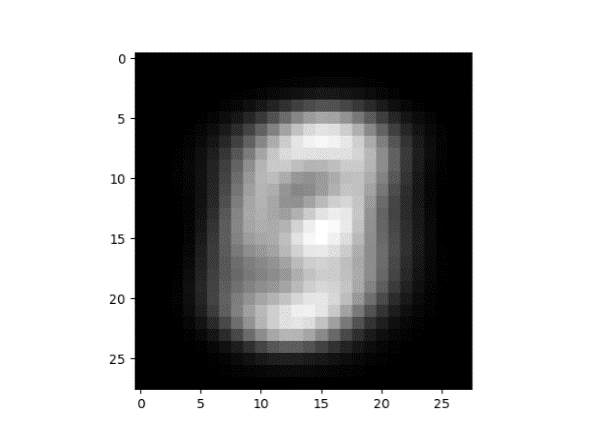}}
		\centerline{(a)}
	\end{minipage}
	\begin{minipage}{0.3\linewidth}
		\centerline{\includegraphics[width=3cm]{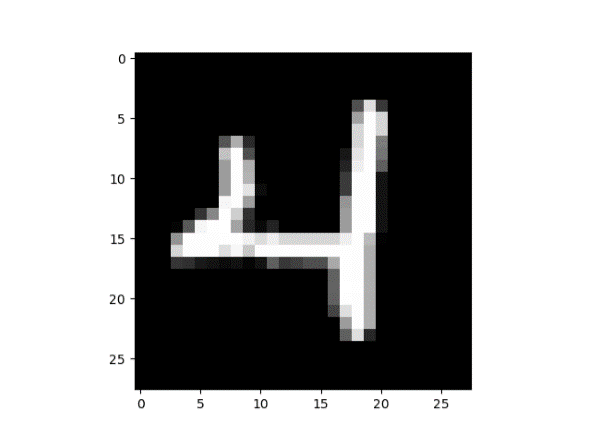}}
		\centerline{(b)}
	\end{minipage}
	\begin{minipage}{0.3\linewidth}
		\centerline{\includegraphics[width=3cm]{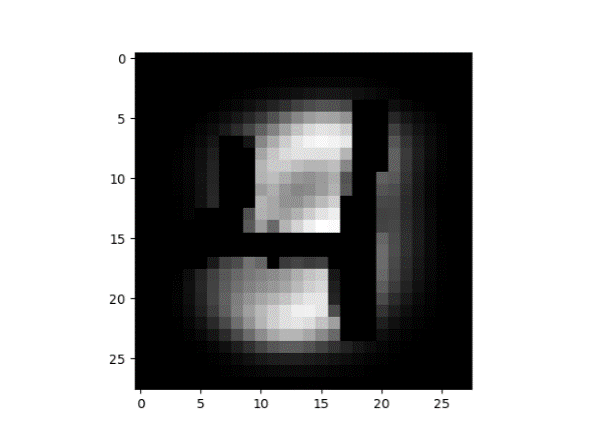}}
		\centerline{(c)}
	\end{minipage}
	\caption{Exclusive Feature Illustration: (a) is the mean feature representation, (b) is one specific example, i.e., digit 4, (c) is the feature representation which exclude this example.} 
	\label{fig1}
\end{figure}

As one key component of unsupervised learning, the autoencoder (AE) is a widely used and promising method aims to learn a latent feature representation of data \cite{baldi2012autoencoders}. It is flexible to implementation and can be stacked easily to form a deep structure, which endows a model the ability to perform complex information extraction and representation \cite{vincent2010stacked}. The AE is with an encoder-decoder structure. Raw data example $x$ is first projected to a latent feature space by the encoder, which is more compressed, sparser, or any specific nature as we need:
\begin{small}
\begin{equation}
\begin{aligned}
h = a_{e}\left(w_{e}\cdot x+b_{e}\right),
\end{aligned}
\end{equation}
\end{small}where $w_{e}$ is the weight of encoder, $b_{e}$ is a bias and $a_{e}$ is a nonlinear activation function used to achieve nonlinearity so that the network can better model complex problem. Next, the latent feature representation $h$ is projected back to the original feature space by the decoder, in which a reconstructed data example $\hat{x}$ is obtained:
\begin{small}
\begin{equation}
\begin{aligned}
\hat{x}=a_{d}\left(w_{d}\cdot h+b_{d}\right),
\end{aligned}
\end{equation}
\end{small}where $w_{d}$ is the weight of decoder, $b_{d}$ is a bias and $a_{d}$ is a nonlinear activation function. The loss of $\hat{x}$ and $x$ is calculated as:
\begin{small}
\begin{equation}
\begin{aligned}
L_a = \left \| x-\hat{x} \right \|_{2}^{2}.
\end{aligned}
\end{equation}
\end{small}The AE learns to minimze Eq. (3) and forces $h$ to retain the most powerful expression of data. 

However, if the AE is endowed with too much capacity, the encoder and decoder may perform among themselves and eventually converge to an identity function \cite{goodfellow2016deep}. In recent years, some related regularization approaches have been proposed to address this issue, such as Denoising AE \cite{vincent2010stacked}, Sparse AE \cite{xu2016stacked}, Graph AE \cite{yu2013embedding}, Winner-take-all AE \cite{makhzani2014winner}, Similarity-aware AE \cite{chu2017stacked}, etc. Despite the efforts made, however, most AE based methods only focus on the reconstruction and may ignore some inherent relation of data, i.e., statistical and geometrical dependence. We argue it an obstacle to learning a more robust and discriminative feature representation. Let us use the handwritten digits as an illustration. In Fig. 1, start from (a), we first obtain the global mean feature representation of whole-set. Then we randomly choose an example, i.e., a digit 4, as shown in (b). Next, we exclude (b) out from the global feature representation and obtain (c). We can observe that the obtained feature representation (c) is radically different from the chosen digit 4. Actually, in an ideal state, if we flatten these two representations, we may obtain two orthogonal vectors. Although the ideal state is not fully available, this exclusivity intuitively exists. Taking the exclusivity into account, we may deal with each example in the case of heterogeneous and homologous respectively, and discover more latent information from the raw data.

This insight inspires us to propose a novel regularization technique to boost the unsupervised feature learning within AE. In this paper, we propose a novel AE based unsupervised feature learning approach called Exclusivity Enhanced Autoencoder (EE-AE), we introduce two exclusivity constraints that can better deal with the statistical and geometrical dependence of data. Moreover, we also make some improvements to the stacked AE structure especially for the connection of different layers on the decoder part, and this makes full utilization of the layer difference and cooperation within AE. Extensive experiments are carried out on several benchmark datasets which show that the proposed EE-AE outperforms other existing representative methods.

\section{Exclusivity Enhanced Autoencoder}

\subsection{Exclusivity Concept}
Data has its own inherent relation, i.e., statistical and geometrical dependence. In unsupervised learning, we don't take labels into account, so how to extract useful information from data itself counts. Suppose we have several examples in total: $C =\left \{ x_1,x_2,\cdots x_n \right \}$, where $C$ is the global-set includes $n$ examples. Let $x_j$ be a random example, then we exclude $x_j$ out from $C$, the rest of global-set is $C^{'} =\left \{ x_1,x_2,\cdots x_n \right \}/\left \{ x_j \right \}$. Then we want to compare $C^{'}$ with $x_j$, normally we need to compare $n-1$ pairs. However, the one to one comparison is not very necessary in this case. So we consider to compare $x_j$ with the global of $C^{'}$, here we replace $C^{'}$ with the mean feature representation $\bar{C^{'}}$. The similarity $S$ between $x_j$ and $\bar{C^{'}}$ is excepted to have a lower bound $\gamma$:
\begin{small}
\begin{equation}
\begin{aligned}
\lim S\left (x_{j}, \bar{C^{'}} \right ) \rightarrow \gamma,
\end{aligned}
\end{equation}
\end{small}where $S$ is a similarity function and $\gamma \rightarrow 0$ under cosine measurement. On the other hand, the rest of global-set $C^{'}$ still contains a certain number of examples that belong to the same class of $x_j$. So we can exclude $m$ examples out from $C^{'}$ and obtains: $D = \left \{ x_1^{m},x_2^{m},\cdots x_m^{m} \right \} \subseteq C^{'}$, where $D$ contains $m\left ( m\ll n \right )$ examples that are most similar with $x_j$. We also replace $D$ with the mean feature representation $\bar{D}$, the similarity $S$ between $x_j$ and $\bar{D}$ is excepted to have a upper bound $\delta$: 
\begin{small}
\begin{equation}
\begin{aligned}
\lim S\left (x_{j}, \bar{D} \right )\rightarrow \delta,
\end{aligned}
\end{equation}
\end{small}where $\delta \rightarrow 1$ under cosine measurement. 

Statistically speaking, the variation between $x_{j}$ and $\bar{C^{'}}$ should be large enough, and the variation between $x_{j}$ and $\bar{D}$ should be small enough. Geometrically speaking, $x_{j}$ and $\bar{C^{'}}$ should be far enough, and $x_{j}$ and $\bar{D}$ should be close enough. We use $\bar{C^{'}}$ to define the heterogeneous case (i.e., different classes) to example $x_j$, and use $\bar{D}$ to define the homologous case (i.e., same class). Although the ideal state is not fully available, i.e., $C^{'}$ contains a certain number of examples whose class is the same as $x_j$, the impact is limited in the global perspective, the exclusivity intuitively exists.

\subsection{Exclusivity Enhanced Autoencoder}
The conventional AE focuses on the reconstruction from the encoder-decoder phase. We consider $n$ examples dataset, similar with Eq. (3), the objective can be written as:
\begin{small}
\begin{equation}
\begin{aligned}
L_a=\sum_{i=1}^{n}\left \| x_i-\hat{x}_i \right \|_{2}^{2}.
\end{aligned}
\end{equation}
\end{small}This objective ignores some inherent relation of data and easily causes overfitting. Inspired by the concept of exclusivity, we find the heterogeneous case $\bar{C^{'(i)}}$ and the homologous case $\bar{D^{(i)}}$ for each example $x_i$ in the original feature space. Then, we focus on the latent feature space and force $h_i$ to apart from $f_e(\bar{C^{'(i)}})$ and close to $f_e(\bar{D^{(i)}})$, respectively. Where $f_e(\cdot)$ is the encoder that projects element to latent feature space.

To this end, we first introduce an auxiliary function $\Omega(\cdot)$ that activates vector $\varepsilon$ on dimension-wise:
\begin{small}
\begin{equation}
\begin{aligned}
\Omega\left(\varepsilon\right) : \left\{ \begin{array}{ll}
\varepsilon_{i} & \textrm{if $\varepsilon_{i} \geq 0$}\\
0 & \textrm{if $\varepsilon_{i} < 0$}
\end{array} \right.,
\end{aligned}
\end{equation}
\end{small}where we activate each dimension $\varepsilon_{i}$ in vector $\varepsilon$. Then, in order to force the latent feature representation of data to apart from their heterogeneous case, we consider a new constraint $L_h^{(1)}$ to maximize the variation between examples and their heterogeneous case:
\begin{tiny}
\begin{equation}
\begin{aligned}
L_h^{(1)} = \sum_{i=1}^{n} \left [ \frac{\Omega \left[f_{e}(\bar{C^{'(i)}})-h_{i}\right]\cdot h_{i}}{\left \| \Omega \left[f_{e}(\bar{C^{'(i)}})-h_{i}\right] \right \|\left \| h_{i} \right \|} \right ] =  \sum_{i=1}^{n} \left [ \frac{\Omega \left[f_{e}(\bar{C^{'(i)}})-f_{e}(x_{i})\right]\cdot f_{e}(x_{i})}{\left \| \Omega \left[f_{e}(\bar{C^{'(i)}})-f_{e}(x_{i})\right] \right \|\left \| f_{e}(x_{i}) \right \|} \right ].
\end{aligned}
\end{equation}
\end{tiny}We can further minimize the variation between examples and their homologous case by considering another constraint $L_h^{(2)}$:
\begin{tiny}
\begin{equation}
\begin{aligned}
L_h^{(2)} =\sum_{i=1}^{n} \left [ \frac{\Omega \left[f_{e}(\bar{D^{(i)}})-h_{i}\right]\cdot h_{i}}{\left \| \Omega \left[f_{e}(\bar{D^{(i)}})-h_{i}\right] \right \|\left \| h_{i} \right \|} \right ]=\sum_{i=1}^{n} \left [\frac{\Omega \left[f_{e}(\bar{D^{(i)}})-f_{e}(x_{i})\right]\cdot f_{e}(x_{i})}{\left \| \Omega \left[f_{e}(\bar{D^{(i)}})-f_{e}(x_{i})\right] \right \|\left \| f_{e}(x_{i}) \right \|} \right ].
\end{aligned}
\end{equation}
\end{tiny}For these two functions, the $L_h^{(1)}$ has a lower bound that close to 0, and the $L_h^{(2)}$ has an upper bound that close to 1. In order to jointly optimize these two terms with conventional AE objective, we subtract $L_h^{(2)}$ from 1, and combine these two terms:
\begin{small}
\begin{equation}
\begin{aligned}
L_h = L_h^{(1)} + \left ( 1 - L_h^{(2)} \right ).
\end{aligned}
\end{equation}
\end{small}Finally, the unified objective of our Exclusivity Enhanced Autoencoder (EE-AE) can be described as: 
\begin{small}
\begin{equation}
\begin{aligned}
L &= L_a + \lambda L_h  \\
&= L_a + \lambda  \left [ L_h^{(1)} + \left ( 1 - L_h^{(2)} \right ) \right ],
\end{aligned}
\end{equation}
\end{small}where we jointly minimize this objective during training. $L_a$ is the conventional AE reconstruction loss, $L_h^{(1)}$ is the exclusivity constraint that maximizes the variation between examples and their heterogeneous case, $(1-L_h^{(2)})$ is the exclusivity constraint that minimizes the variation between examples and their homologous case. $\lambda$ is an hyper-parameter controls the balance between the conventional reconstruction loss and added constraints.

\subsection{Stacked Autoencoders Structure}
The conventional stacked autoencoders (AEs) adopt the stacked structure to every single AE \cite{vincent2010stacked,maria2016stacked}. Each time, the latent feature representation $h$ of the current AE is fed to the input of next AE. Such a process is repeated with certain steps to realize the stacked structure.

Recently, Yosinski et al. \cite{yosinski2014transferable} experimentally quantified the generality versus specificity of each layer in DNNs, and shown that the feature representation of an example normally go through a {\bf specific} $\rightarrow$ {\bf general} $\rightarrow$ {\bf specific} three level phases from the input to output layers, the deeper the more general. Inspired by this analysis, we propose to fully consider the connection of different layers from decoders. Specifically, we not only stack the encoder but also the decoder. Then in the last stacked structure network, we further fine-tune the whole network to achieve better cooperation of different AE layers (Fig. 2). This can be regarded as a weight initialization trial. we constrain the weight variance ratio of layers as: $1-\eta \leq \frac{\left \| W_F \right \|_{p.1}}{\left \| {W_F}' \right \|_{p.1}} \leq 1+\eta$, where $W_F$ and ${W_F}'$ are the original and fine-tuned weight respectively. $\left \| \cdot  \right \|_{p.1} $ is the $l_p$ norm on a vector and $\eta$ is a hyper-parameter controls the weight variance ratio of layers.

\begin{figure}[htbp]
	\centering
	\includegraphics[width=0.35\textwidth]{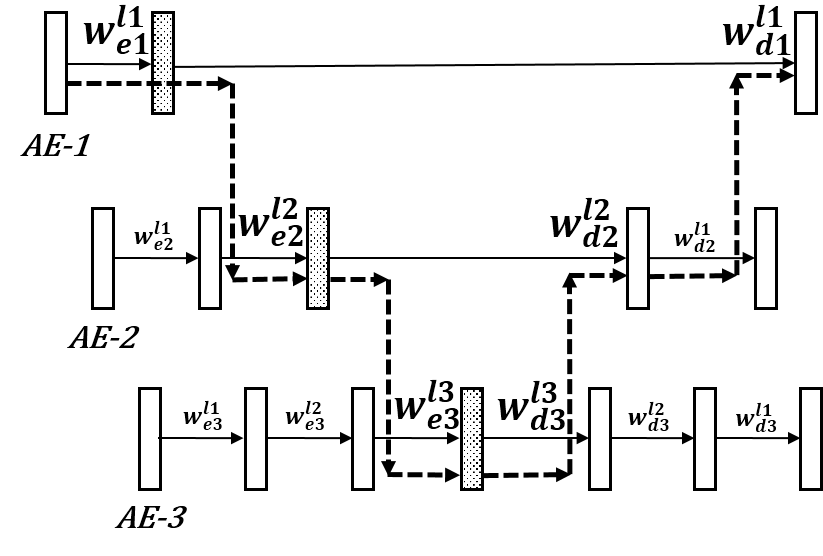}    
	\caption{Modified Stacked Autoencoders Structure: A case for 3 AEs (AE-1 to 3). Solid line stands for each single AE and the dotted line is the stacked structure. $W_{ei}^{li}$ and $W_{di}^{li}$ is the weight for encoder and decoder, respectively.}
	\label{Fig2}
\end{figure}

\section{Experiments}
We compare our proposed Exclusivity Enhanced Autoencoder (EE-AE) with several state-of-the-art methods on two widely used datasets COIL100 \cite{nene1996columbia} and MNIST \cite{lecun1998gradient}. The experimental results and analysis are discussed in detail for each dataset respectively.

\subsection{Results on COIL100}

\textbf{Experiment Setup:} The COIL100 contains 7200 color images of 100 objects. Images of the objects are taken at pose intervals of 5 degrees, corresponding to 72 poses per object. We convert these images to grayscale images and resize them to 32x32 pixels. We randomly select 10 images for each object to form the training set and the rest images are the testing set. For the training set, we also consider their horizontal mirror feature. We use our proposed EE-AE for unsupervised feature learning. The nearest neighbor classifier whose inputs are the learned feature representations from EE-AE is applied to achieve the recognition. These processes are repeated 10 times and we report the average recognition results. We also split the dataset to 20, 40, 60, 80 object subsets and compare the recognition accuracy on these subsets and whole-set respectively. We compare our proposed approach with 9 existing representative methods (Table 1). All competitors and our EE-AE are under the same settings and tuned to achieve their best performance. For all AE based methods, we also adopt the same network architecture and report the results on the learned feature representations of the last stacked hidden embedding layer.
\begin{table}[htbp]
	\begin{center}
		\caption{Accuracy (\%) for COIL100 }
		\setlength{\tabcolsep}{0.8mm}{ 
			\begin{tabular}{|l|c|c|c|c|c|c|}        
				\hline                   
				Method         &20  &40  &60  &80  &100  &Average\\
				\hline
				PCA \cite{turk1991eigenfaces}           &90.41      &89.27      &87.26      &85.80      &84.17       &87.37 \\
				\hline
				KPCA \cite{scholkopf1998nonlinear}          &90.12      &88.06      &85.75      &83.97      &82.52       &86.08 \\  
				\hline
				NPE \cite{he2005neighborhood}           &91.76      &89.59      &87.27      &85.94      &85.03       &87.91 \\
				\hline
				SCC \cite{cai2011sparse}           &91.49      &89.39      &85.91      &83.91      &82.91       &86.72 \\  
				\hline                                                        
				SDNMF \cite{qian2016non}         &89.91      &87.88      &82.94      &81.44      &78.60       &84.15 \\  
				\hline                                                            
				Denoise-AE \cite{vincent2010stacked}    &91.75      &90.52      &88.58      &86.35      &85.17       &88.47 \\  
				\hline                                                               
				Sparse-AE \cite{xu2016stacked}     &92.26      &91.17      &88.59      &86.63      &85.55       &88.83 \\ 
				\hline                                                              
				Graph-AE \cite{yu2013embedding}      &93.37      &91.33      &89.11      &86.67      &85.96       &89.28 \\ 
				\hline                                                              
				SSA-AE \cite{chu2017stacked}        &96.97      &94.04      &91.87      &90.42      &88.78       &92.41 \\      
				\hline                                                           
	            {\bf EE-AE (ours)}   &{\bf 98.39}&{\bf 95.20}&{\bf 93.09}&{\bf 92.58}&{\bf 90.90} &{\bf 94.03}\\
				\hline  
		\end{tabular}
		}
	\end{center}  
\end{table}

\textbf{Results \& Analysis:} The results are shown in Table 1. We can see that our proposed EE-AE outperforms all competitors with great advantage on each subset contains 20, 40, 60 and 80 objects, and whole-set (100 objects) respectively. What's more, we can also have the observation that all the AE based methods generally obtain better performance than traditional methods, which highlights the merit of AE for unsupervised feature learning. Then we further evaluate the sensitivity of the hyper-parameters: (1) The $\lambda$ controls the balance between the conventional reconstruction loss and exclusivity constraints in Eq. (11). (2) The number ($m$) of most similar examples when finding the homologous case ($\bar{D}$) for each example; (3) The number of stacked AEs (denote as $s$ for simplicity). The evaluation is conducted on the 20-object subset of COIL100. When evaluating the sensitivity of one hyper-parameter, we fix others to their best points. Fig. 3 shows the results. We can see that as $\lambda$ becomes larger, the performance of our model becomes better and then decreases, the optimal $\lambda$ is around 7. As to $m$, as it becomes larger, the performance of our model becomes better and finally tends to be relatively stable, and the optimal $m$ is around 6. As to $s$, the optimal number is 3.

\begin{figure}[h]
	\centering
	\begin{minipage}{0.77\linewidth}
		\centerline{\includegraphics[width=6cm]{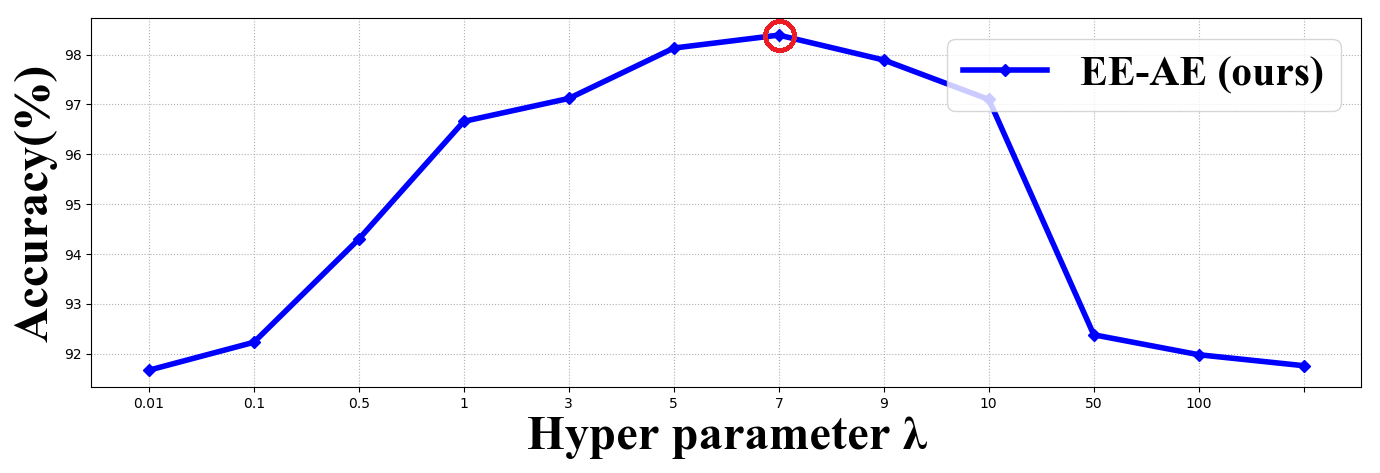}}
	\end{minipage}
	\begin{minipage}{0.77\linewidth}
		\centerline{\includegraphics[width=6cm]{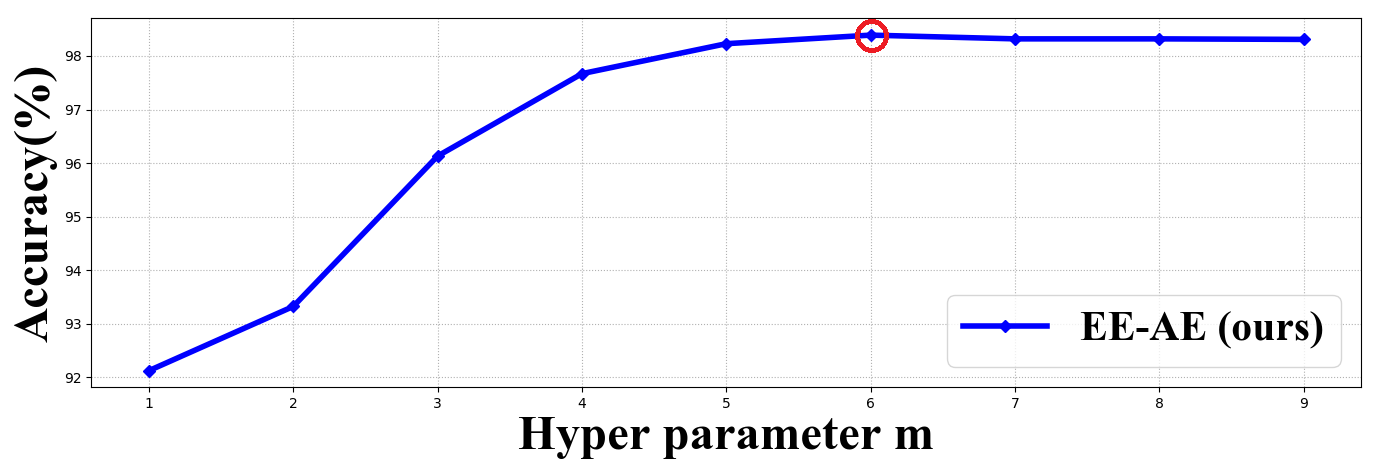}}
	\end{minipage}
	\begin{minipage}{0.77\linewidth}
		\centerline{\includegraphics[width=6cm]{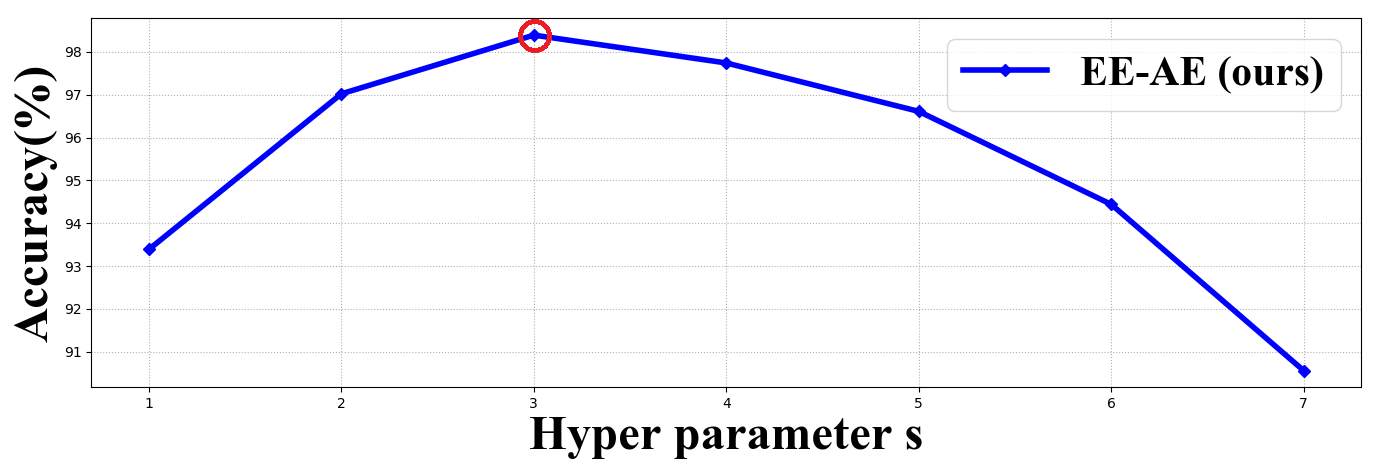}}
	\end{minipage}
	\caption{Sensitivity of Hyper-parameters} 
	\label{fig3}
\end{figure}

\subsection{Results on MNIST}

\textbf{Experiment Setup:} The MNIST \cite{lecun1998gradient} contains 60,000 training images and 10,000 testing images. In order to better evaluate the learning ability of our proposed EE-AE, we split the MNIST into two scenarios: (1) $1$-set: The whole-set. (2) $\frac{1}{6}$-set: Only contains 10000 (out of 60000) training images and 10,000 testing images. We train EE-AE and classifier under these two scenarios respectively. Our proposed EE-AE is used for unsupervised feature learning. With the learned feature representations, we train an SVM classifier to achieve the recognition. As to the SVM, we adopt 1-versus-1 setting \cite{duan2007one,guo2016improved} for the multi-class classification. The purpose of our experiments on the MNIST is to quantitatively evaluate the effectiveness of our model.

\textbf{Results \& Analysis:} The results are shown in Table 2. When evaluating on $1$-set scenario, our model obtains a comparable performance with these competitors. However, when evaluating on $\frac{1}{6}$-set, our model outperforms these competitors with great advantages. This shows that our proposed EE-AE has significant better learning ability to discover and extract useful information from data, even the data volume is relatively small. This results in a better generalization ability.
%
%

\begin{table}[t]
\begin{center}
\caption{Accuracy (\%) for MNIST}
\setlength{\tabcolsep}{3.1mm}{ 
\begin{tabular}{|l|l|c|c|}
\hline
\multicolumn{2}{|l|}{Method}                  & $1$-set & $\frac{1}{6}$-set \\ \hline
\multicolumn{2}{|l|}{SAE \cite{vincent2010stacked}}  & 98.60   & 90.37   \\ \hline
\multicolumn{2}{|l|}{SDAE \cite{vincent2010stacked}} & 98.72   & 91.04   \\ \hline
\multirow{6}{*}{\textbf{EE-AE (ours)}} & $\eta=0.0$        & 98.43   & 93.56   \\ \cline{2-4} 
                              & $\eta=0.2$           & 98.59   & 94.33   \\ \cline{2-4} 
                              & $\eta=0.4$           & 98.67   & 94.89   \\ \cline{2-4} 
                              & $\eta=0.6$           & 98.67   & \textbf{95.23}   \\ \cline{2-4} 
                              & $\eta=0.8$           & \textbf{98.81}   & 94.61   \\ \cline{2-4} 
                              & $\eta=1.0$             & 98.73   & 94.57   \\ \hline
\end{tabular}
}
\end{center} 
\end{table}

\subsection{Environment and Implementation}
We use Pytorch to implement our experiment. As to AEs, we apply convolution, batch normalization, ReLU and maxpooling layers to form the encoder. The convolution layers have 3 settings: 16 channels, 3x3 kernel, stride=1 and padding=1. 32 channels, 3x3 kernel, stride=1 and padding=0; 64 channels, 3x3 kernel, stride=1 and padding=0. For all the maxpooling layers, we set a 2x2 kernel. As to the decoder, we apply de-convolutional layers which are exactly the reverse of convolution to obtain up-exampled feature map. Also, we use batch normalization, ReLU and maxunpooling layers to cooperate with the network. We append an embedding layer on the top of the hidden codes with 128-dimensional features. The stochastic gradient descent (SGD) is applied for the optimization.

\section{Conclusion and Future Work}
In this paper, we propose a novel autoencoder based unsupervised feature learning approach called Exclusivity Enhanced Autoencoder (EE-AE). We utilize the exclusivity constraints to cooperate with feature extraction within AE. Our model can better deal with the statistical and geometrical dependence of data and results in a more robust and discriminative feature representation. Extensive experiments verified the effectiveness of our model. In the future, we plan to investigate more efficient techniques and apply the exclusivity to larger datasets.

\bibliographystyle{IEEEbib}
\bibliography{my}
\end{document}